\documentclass[conference]{IEEEtran}
\IEEEoverridecommandlockouts
\usepackage{cite}
\usepackage{amsmath,amssymb,amsfonts}
\usepackage{algorithmic}
\usepackage{graphicx}
\usepackage{textcomp}
\usepackage{xcolor}
\usepackage{hyperref}
\usepackage{subfig}
\usepackage{booktabs}
\usepackage{multirow}
\usepackage{pifont} %\ding{55}, \ding{51}
\def\BibTeX{{\rm B\kern-.05em{\sc i\kern-.025em b}\kern-.08em
    T\kern-.1667em\lower.7ex\hbox{E}\kern-.125emX}}
\begin{document}

\title{Deep Event-based Object Detection in Autonomous Driving: A Survey}

\author{\IEEEauthorblockN{1\textsuperscript{st} Bingquan Zhou}
\IEEEauthorblockA{\textit{School of System Engineering} \\
\textit{NUDT}\\
Changsha, China \\
bingquanzhou@nudt.edu.cn}
\and
\IEEEauthorblockN{2\textsuperscript{nd} Jie Jiang*}
\IEEEauthorblockA{\textit{School of System Engineering} \\
\textit{NUDT}\\
Changsha, China \\
jiejiang@nudt.edu.cn}
\thanks{* Corresponding author: jiejiang@nudt.edu.cn.}
}

\maketitle

\begin{abstract}
Object detection plays a critical role in  autonomous driving, where accurately and efficiently detecting objects  in fast-moving scenes is crucial. Traditional frame-based cameras face  challenges in balancing latency and bandwidth, necessitating the need  for innovative solutions. Event cameras have emerged as promising  sensors for autonomous driving due to their low latency, high dynamic  range, and low power consumption. However, effectively utilizing the  asynchronous and sparse event data presents challenges, particularly in  maintaining low latency and lightweight architectures for object  detection. This paper provides an overview of object detection using  event data in autonomous driving, showcasing the competitive benefits of event cameras. 
\end{abstract}

\begin{IEEEkeywords}
Event-Based Camera, Object Detection, Autonomous Driving,  Neuromorphic Vision, Spiking Neural Network, Graph Neural Network, Multi-modal Learning, Deep Neural Network
\end{IEEEkeywords}

\section{Introduction}
Time plays a critical role in object detection, particularly in fast-moving vehicles. Traditional frame-based cameras face a trade-off between latency and bandwidth. This has prompted the development of event cameras, which offer low latency (in the order of microseconds) and high dynamic range (140 dB compared to 60 dB in frame-based cameras)\cite{Gallego_2019_event_survey}. Moreover, event cameras exhibit high temporal resolution and consume low power, making them competitive sensors for autonomous driving. However, effectively handling the asynchronous and sparse event data poses challenges for researchers. Various techniques have been employed, including utilizing conventional deep neural networks (DNN) architecture\cite{Iacono_2018_offshelf}\cite{Chen_2018_Pseudo}\cite{Cannici_2019_YOLE}\cite{Messikommer_2020_asynet}\cite{de_Tournemire_2020_red}\cite{Xu_2021_STMLP}\cite{Li_2022_astmnet}\cite{Liu_2023_FRLW}\cite{Wang_2023_dual}\cite{Gehrig_2023_rvt}, leveraging bio-inspired Spiking Neural Networks (SNN)\cite{Kugele_2021_hybrid}\cite{Cordone_2022_spike_ssd}\cite{Zhang_2023_FPDAG}\cite{Nagaraj_2023_DOTIE}, learning spatio-temporal features using Graph Neural Networks (GNN)\cite{Schaefer_2022_GNN}\cite{Gehrig_2022_EAGR}, and combining data fusion with RGB frames\cite{Jiang_2019_mixed}\cite{Li_2019_JDF}\cite{Liu_2021_ADF}\cite{Tomy_2022_fusionFPN}\cite{Zhou_2023_RENet}\cite{Li_2023_sod}. Additionally, to enhance research in object detection with event data, several datasets have been proposed\cite{de_Tournemire_2020_red}\cite{Perot_2020_dataset}\cite{Zhou_2023_RENet}\cite{Li_2023_sod}.

\begin{figure}[!htbp]
	\centering
	\includegraphics[width=0.45\textwidth]{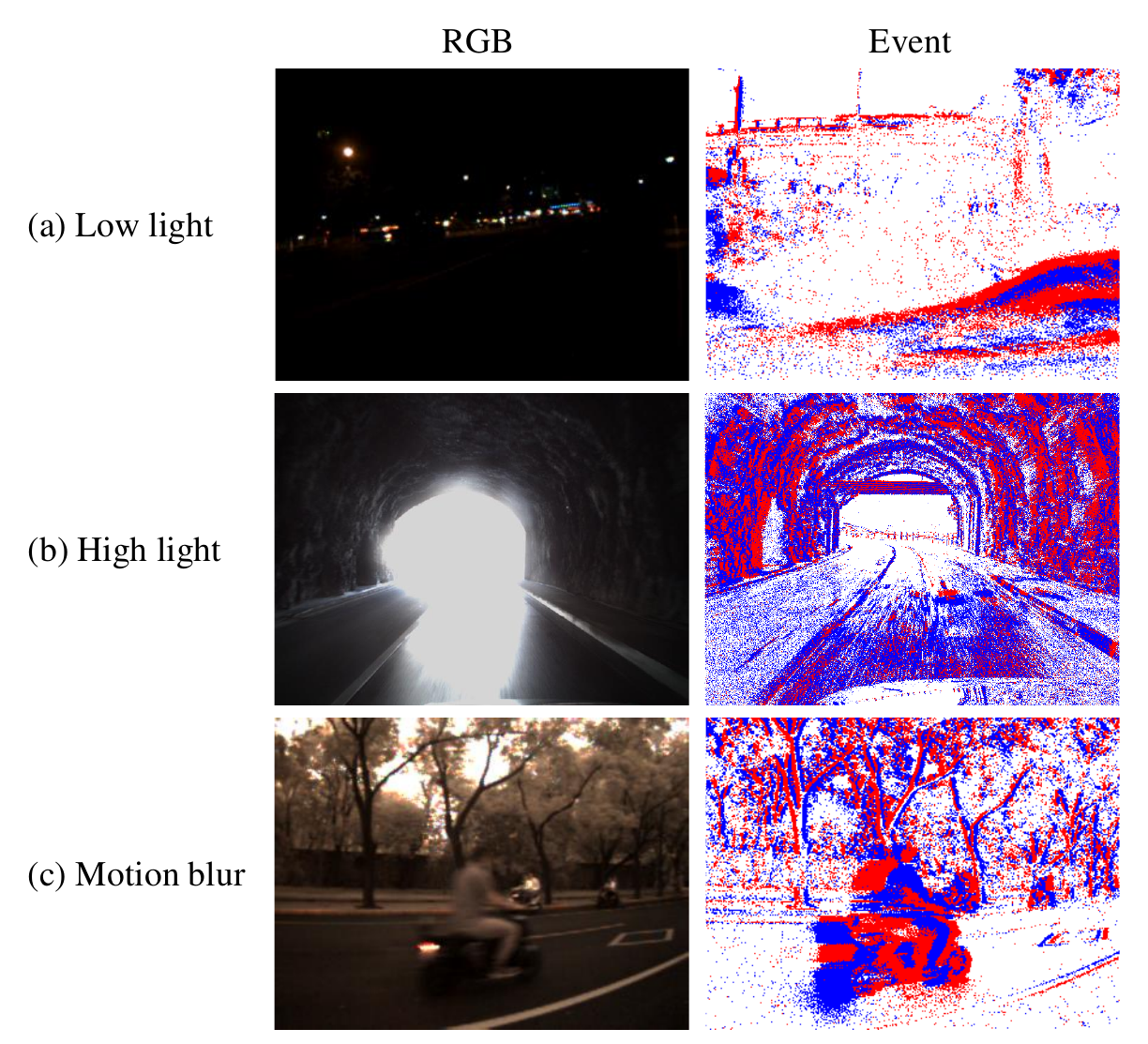}
	\caption{Comparison between RGB and Event Cameras. (a) In low-light conditions, RGB cameras fail to distinguish objects from the background due to insufficient light capture, whereas event cameras excel at detecting object edges. (b) RGB cameras' limited dynamic range falls short in high-intensity scenes, unlike event cameras with their high dynamic range for clear object detection. (c) Motion blur in RGB imaging during fast movements is avoided with event cameras, which offer low-latency, blur-free detection.}
	\label{fig:1}
\end{figure}

In this paper, we present an overview of the object detection process using event data and introduce cutting-edge research in this domain. The subsequent sections are organized as follows: Section 2 covers the event camera, event data structure, and the object detection task. Section 3 discusses several event preprocessing methodologies. Section 4 presents recent event-based object detection techniques employing different types of neural networks. Section 5 reviews datasets tailored for object detection using event cameras. Finally, Section 6 concludes with a discussion. 

\section{Event Camera and Event Data}
\subsubsection{Event camera} Event camera (Neuromorphic camera) is a bio-inspired camera in which each photoreceptor
 behind a pixel mimics a light-sensitive cell in biological retinas, as shown in Figure 1. When the illumination changes, each pixel on the event camera will record the log intensity, and send a spike(event) once the change exceeds a set threshold\cite{Gallego_2019_event_survey}. The spike signals are transmitted from the retina pixel and then out of the camera, using a key technique \textbf{AER}\cite{Liu_2015_aer} to map the address. Thus, the event cameras can asynchronously output the event stream. The first event camera is Dynamic Vision Sensor  (\textbf{DVS}), which was developed for frame-free and event-driven tasks\cite {Lichtsteiner_2006_dvs}. The Asynchronous Time-Based Image Sensor (\textbf{ATIS})\cite{Posch_2010_atis} succeeds the component in DVS and makes it possible to output the absolute intensity in illumination change. The ATIS achieves a very large static dynamic range (\textgreater120 dB) but still have some disadvantage (For example, the pixel is at least double the area of DVS pixels\cite{Gallego_2019_event_survey}). The most usable event camera called \textit{Dynamic and Active Pixel Vision Sensor} (\textbf{DAVIS})\cite{Berner_2013_davis} combines the DVS and a conventional active pixel sensor (\textbf{APS}), which outputs grayscale video at limited dynamic range(55 dB). This combination makes it possible to use both the information from event-driven and frame-based cameras. 
 \begin{figure}[!htbp]
	\centering
	\includegraphics[width=0.45\textwidth]{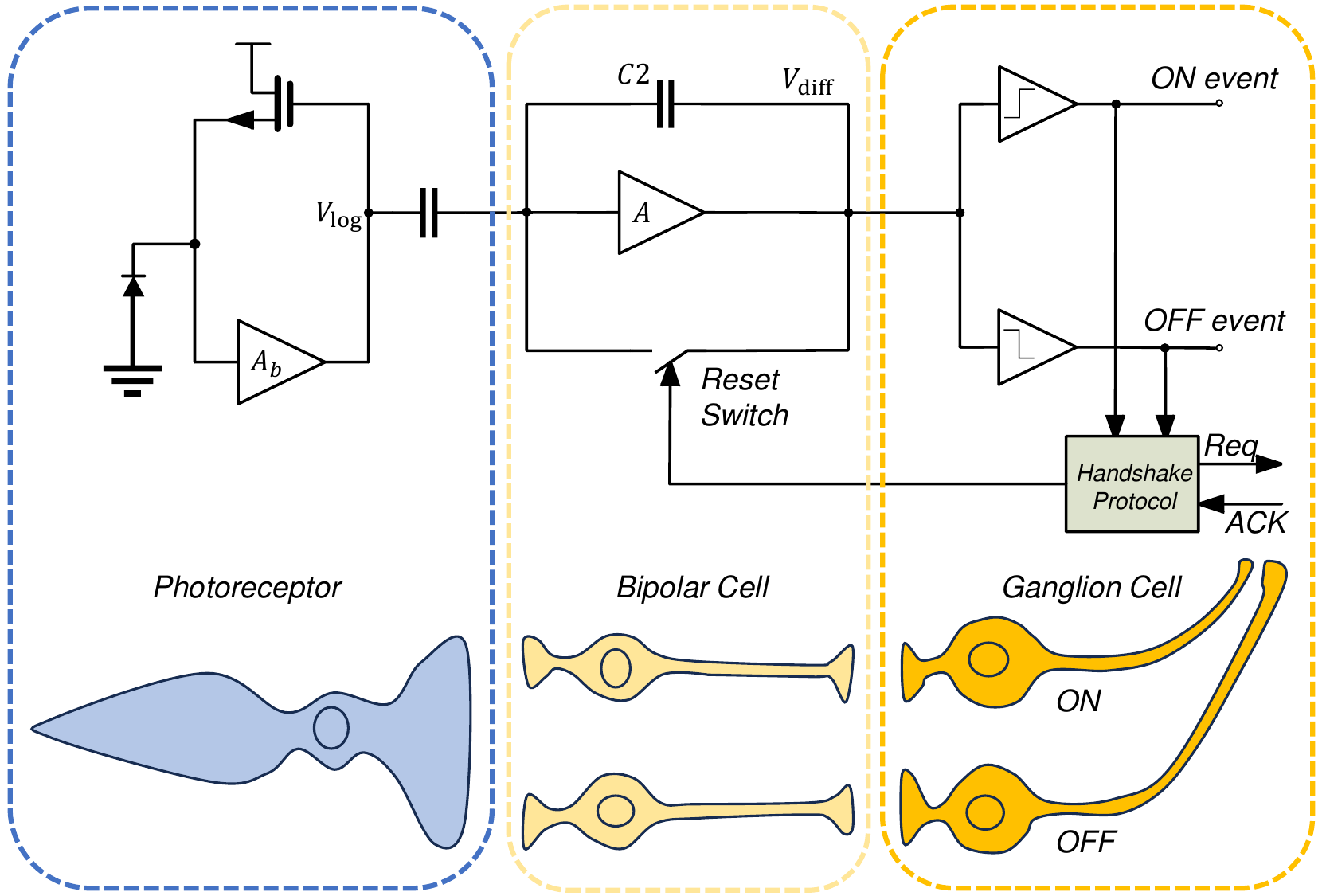}
	\caption{Comparative diagram of event camera circuitry and visual nerves(adapted from \cite{event_hardware}). Here, the circuit diagram mimics three parts of the vision sensing nerve: the photoreceptors are responsible for converting signals into electrochemical signals that can be transmitted by neurons, the ganglion cells receive the electrochemical signals, and then these electrochemical signals continue to be transmitted along the optic nerve, ultimately becoming a binary signal.}
	\label{fig:2}
\end{figure}

 \subsubsection{Event Data Structure} Due to its information perception mechanism, the camera outputs event data as a time-related type. Specifically, the event data is structured as an (x,y,p,t) tuple, where (x,y) indicates the pixel's position; p represents the light-changing polarity; and t is the time stamp. Polarity signifies the change in lighting intensity, corresponding to ON and OFF events. Therefore, from a spatio-temporal dimension perspective, event data consists of a collection of discrete points in space and time, often containing rich spatial and temporal semantic information, which aids in tasks highly related to space and time, such as real-time object detection in autonomous driving scenarios.

\section{Event Data Preprocessing}
Unlike synchronous and dense images in the 2D vision tasks, the event data can't be directly processed by conventional computer vision methods due to its asynchronous and sparse form\cite{Messikommer_2020_asynet}. Thus, several methods are proposed to pre-process the event data to bring the advanced performance of the vision deep learning to the event-based vision tasks.

\subsubsection{Event Frame (Event Image)} is a simple way to convert the event data to a structure that traditional DNN can deal with, which has been used in several event-based object detection methods\cite{Iacono_2018_offshelf}\cite{Jiang_2019_mixed}\cite{Liu_2021_ADF}\cite{Tomy_2022_fusionFPN}\cite{Li_2023_sod}. It compressed the sparse and asynchronous event point in the spatio-temporal space into many synchronous and dense image-like frames. The event frame can be understood as a type of 2-channel image, composed by positive events and negative events. Such representation can provide dense features which can well suit in modern deep learning methods\cite{Gehrig_2023_rvt}, since the spatial information of the object edges, which can be sensed by events, is the most informative to 2D computer vision algorithms. However, it can also discard the sparsity of the event data, leading to the raising of  memory usage and computation.

\subsubsection{Voxel Grid} is adapted from the method dealing with point clouds. It understands the spatio-temporal tuple $(x,y,t)$ as the $(x,y,z)$ in 3D space. Just like a pixel in 2D space, a event voxel is a fixed voxel in spatio-temporal dimension, which is given by a specific pixel and time interval\cite{Zhu_2019_voxel}. Compared to evnet frame, this representation contains more temporal information by avoiding squeeze the time dimension on a 2D gird\cite{Gallego_2019_event_survey}. For a set of events within a time window $\Delta T$, a $B \times H \times W$ voxel grid can be given as follows:

\begin{align}
    V(x,y,t)&=\sum\limits_{i}p_i\delta_b(x-x_i)\delta_b(y-y_i)\delta_b(t-t_i^*)\\
    t_i^*&=\frac{B-1}{\Delta T}(t_i-t_1)\\
    \delta_b(a)&=max(0,1-|a|)
\end{align}

where B is the time interval, H is the height, and W is the width of the voxel. For voxel grid is still a dense representation somehow, it will also cause the raise of memory and computation consumption.

\subsubsection{Learnable Representation} is a different way in creating event representations. Unlike traditional handcrafted discriptor of events, learnable event representation aims at finding the most suitable one for different scenarios, especially for object detection. For example, EventPillar\cite{Wang_2023_dual}, which takes PointPillar\cite{Lang_2019_pointpillar} as a reference, is optimized through the training process, converting both positive and negative polarity of events into a set of pillars, and generates a 3D tensor representation.  

\begin{figure}[!htbp]
	\centering
	\includegraphics[width=0.5\textwidth]{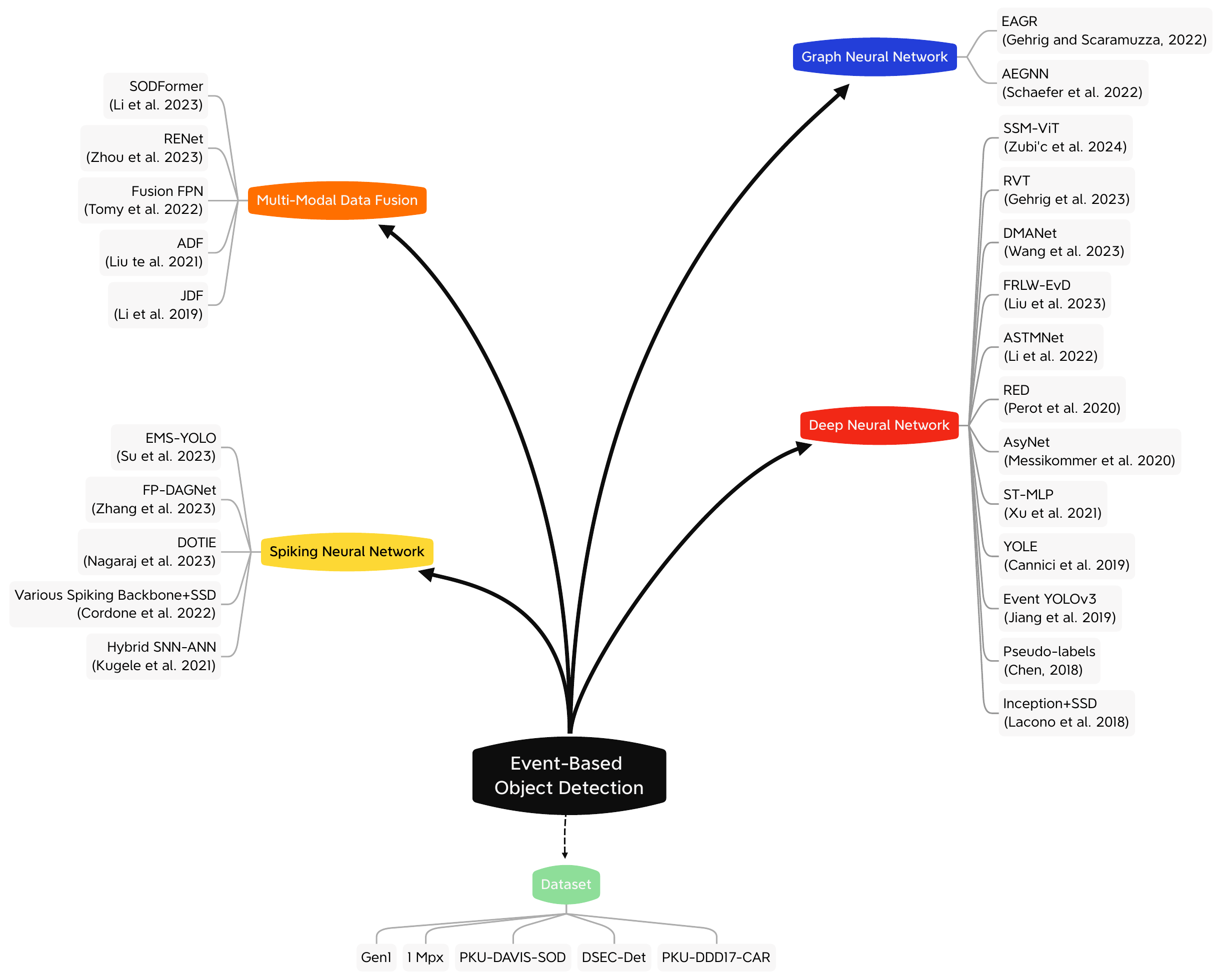}
	\caption{Summary diagram of object detection technology for autonomous driving scenarios based on event cameras and deep learning. It can be broadly divided into four technical approaches: based on traditional deep neural networks, based on graph neural networks, based on spiking neural networks, and based on multi-modal fusion. At the same time, large object detection datasets based on event cameras in driving scenarios is also proposed to support these technologies.}
	\label{fig:1}
\end{figure}

\section{Methodology for Event-Based Object Detection}
The structure of event-based data significantly diverges from that of traditional frame-based imagery, rendering object detection and tracking tasks particularly challenging. In recent years, a variety of approaches have been developed to address these challenges. These methods range from adapting Artificial Neural Networks (ANNs) to the asynchronous nature of event streams, employing architectures that align with the characteristics of events from neuromorphic and graph-based perspectives, to integrating additional information from frame-based images. These strategies have demonstrated success in object detection tasks.

\begin{figure}[!htbp]
	\centering
	\subfloat[Event-Based Object Detection with DNN]{
		\includegraphics[width=0.4\textwidth]{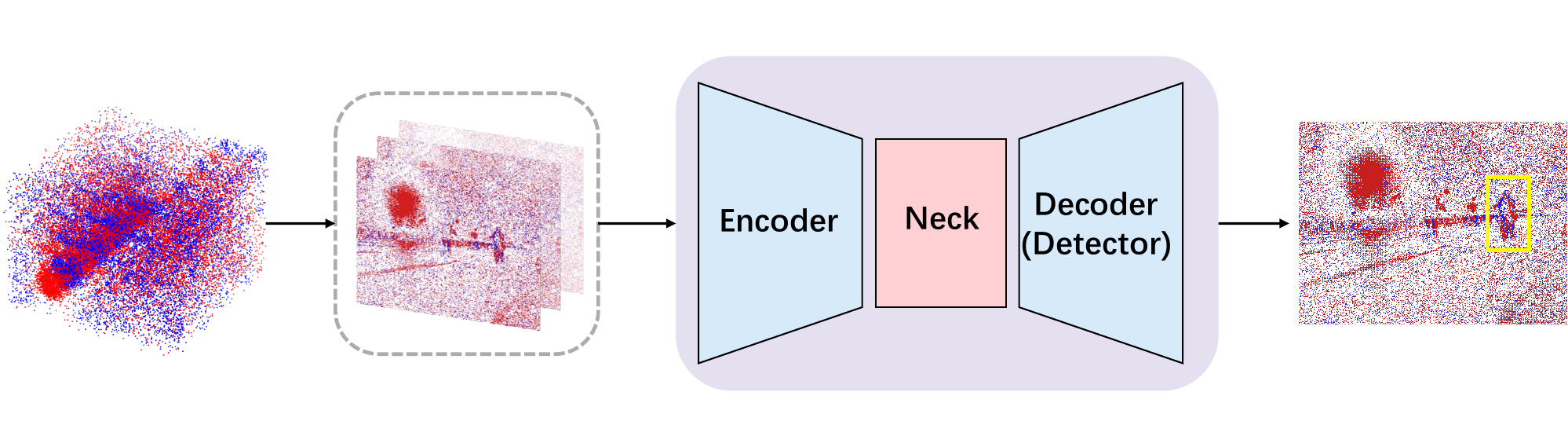}}\\
	\subfloat[Event-Based Object Detection with GNN]{
		\includegraphics[width=0.4\textwidth]{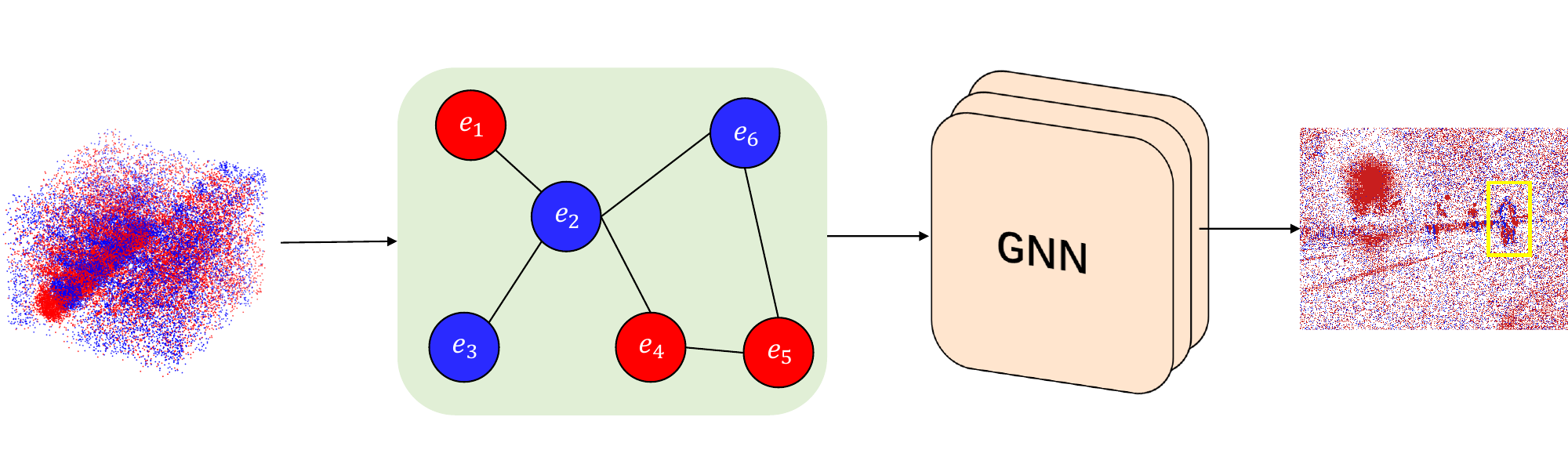}}\\
	\subfloat[Event-Based Object Detection with SNN]{
		\includegraphics[width=0.4\textwidth]{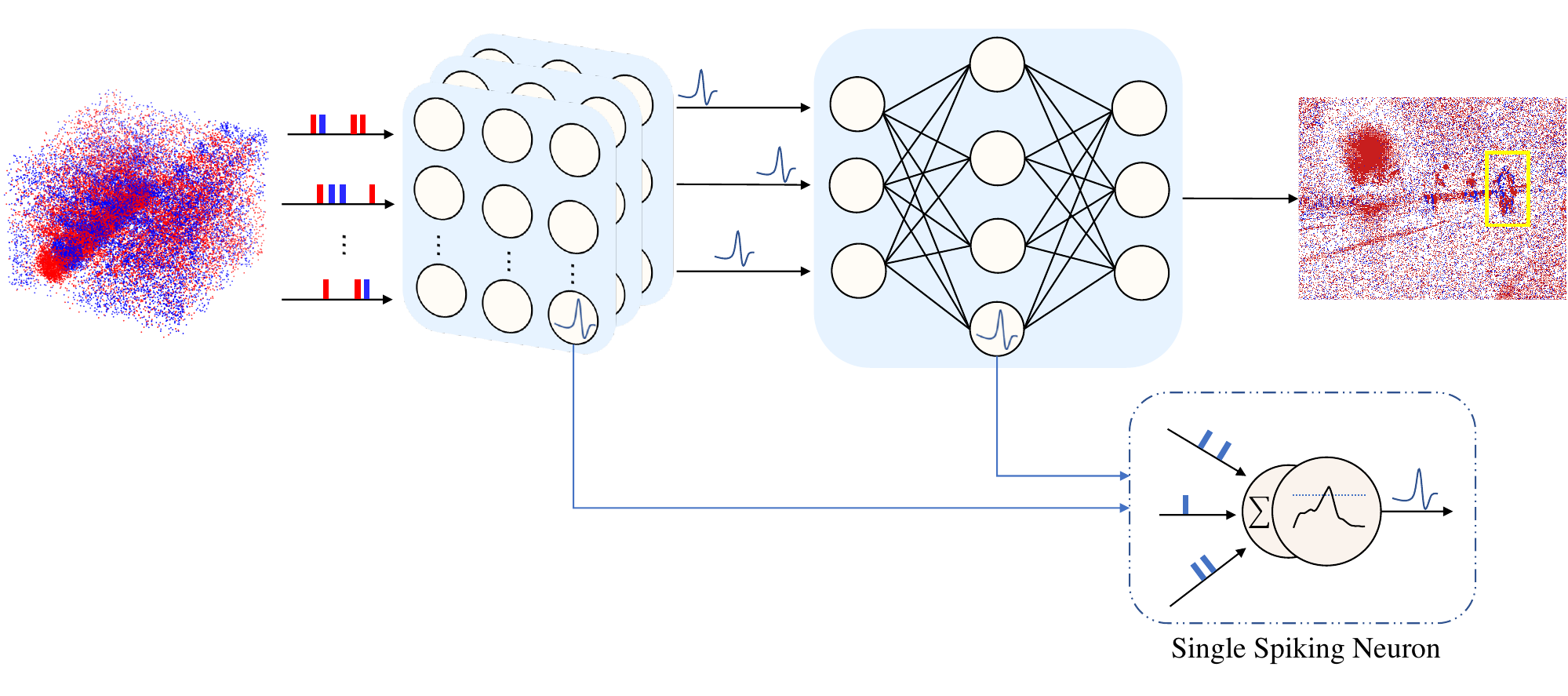}}\\
	\subfloat[Event-Based Object Detection with Mullti-Modal Fusion]{
		\includegraphics[width=0.4\textwidth]{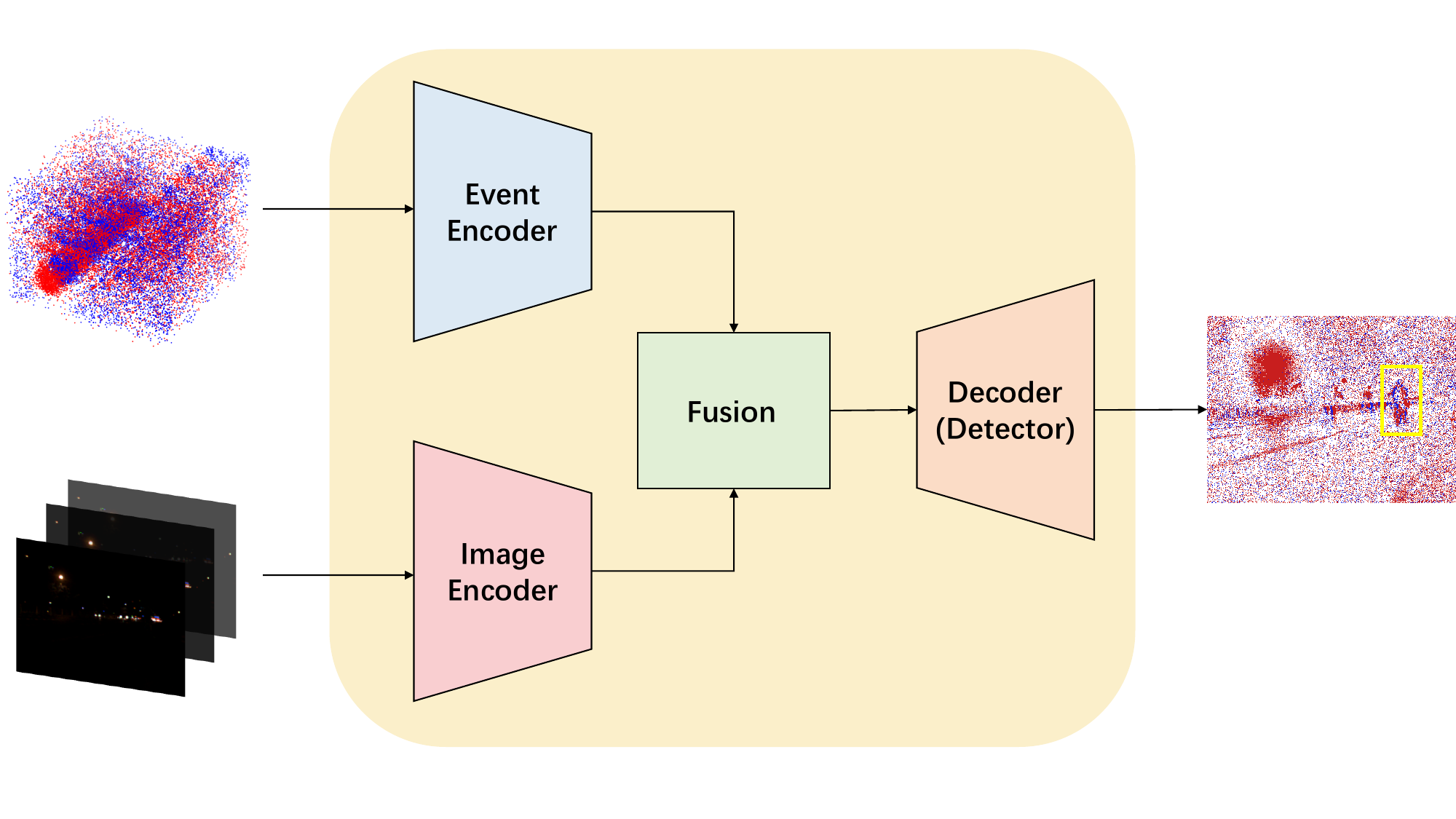}}
	\caption{Different Method in Event-Based Object Detection. (a) Deep Neural Networks (DNNs) for object detection in event data are often applied after transforming raw events into an image-like format, facilitating the use of established models like YOLO. (b) Exploiting the spatio-temporal content of event data is enhanced by modeling it as a graph, rather than compressing it into frames, making it compatible with Graph Neural Networks (GNNs). (c) Spiking Neural Networks (SNNs) directly process asynchronous event data, where individual spiking neurons generate asynchronous spike outputs once the number of received spikes reaches a threshold.  (d) Employing a multimodal fusion strategy, this approach capitalizes on the synergistic attributes of event data and RGB imagery. Feature extraction is initially performed separately on each modality using respective feature extractors, followed by a fusion process. The integrated features are then decoded to determine the bounding boxes and classify the objects. }
	\label{fig_6}
\end{figure}

\subsection{DNN for Event-Based Object Detection}
\subsubsection{YOLO Based Method} 
A common idea to detect object on events is to adapt advanced architecture used in frame-based vision such as YOLO\cite{Bochkovskiy_2020_yolov4} and DETR\cite{Carion_2020_detr}\cite{zhu2021deformable}. To make the event data suitable for object detector, using the event frame representation can be a direct way. For example, \cite{Chen_2018_Pseudo} uses Recurrent Rolling Convolution (RPC)\cite{Ren_2017_RPC} to generate pseudo labels from the APS frame, which is produced by DAVIS\cite{Berner_2013_davis} along with event streams, and use the label and ground-truth events to train a YOLO network. 

However, most of the aforementioned methods preprocess events into event frames, allowing them to directly utilize existing neural network modules for processing. While this approach facilitates the utilization of neural networks, it also sacrifices the inherent sparsity and spatio-temporal characteristics of event data. Therefore, designing novel network architectures that can handle sparse event points in the spatio-temporal domain has become an important research direction. Cannici\cite{Cannici_2019_YOLE} has redefined convolution and pooling operations to alter the forward propagation of fully convolutional networks, recomputing only the feature maps corresponding to the regions affected by events. 
\subsubsection{LSTM Based Method} 
Due to the spatio-temporal characteristics of event streams, the use of Long Short-Term Memory (LSTM) networks\cite{LSTM} to learn features in the temporal domain enables the network to better capture spatio-temporal features. DMANet \cite{Wang_2023_dual}utilizes the hidden state of adaptive convolutional LSTM to encode long-term memory and models short-term memory by computing the spatio-temporal correlations between event pillars within adjacent time intervals. Furthermore, for the task of object detection and the TAF (Temporal Active Focus) event representation, Liu\cite{Liu_2023_FRLW} designed a new module called BFM (Bifurcated Folding Module) and a lightweight object detection head called AED (Agile Event Detector) to better match the high temporal resolution of event data. AED combines the outputs of TAF and BFM, providing competitive accuracy while maintaining high speed.
\subsubsection{Attention Based Method} 
After the self-attention structure\cite{visionTransformer} has shown great promise in the visual domain, the Transformer structure has also been introduced into the field of event data. Gehrig proposed the RVT module\cite{Gehrig_2023_rvt}, which utilizes convolutional priors and local/dilated global self-attention through a multi-stage design to achieve spatial feature interaction. By aggregating recurrent temporal features, RVT achieves minimized inference while preserving event information. SSM-ViT takes a further step by introducing state-space models (SSMs) for event cameras, aiming to address the challenge of poor generalizability in deep neural networks processing event-camera data\cite{Zubi’c_2024_ssm}. SSMs with learnable timescale parameters adapt to varying frequencies without requiring retraining at different frequencies\cite{gu2022efficiently}. Different from RVT, Li\cite{Li_2022_astmnet} proposed the use of a temporal attention convolution module and an adaptive temporal sampling strategy to learn asynchronous attention embeddings from continuous event streams. This asynchronous attention embedding captures key motion information from event data, which is beneficial for continuous object detection.

\subsubsection{Point Cloud Inspired Method} 
Due to the sparsity of event data in the spatial-temporal dimensions and the sparsity of point cloud data in three-dimensional space, many deep learning methods for event cameras draw inspiration from the approaches used in 3D vision for processing point clouds. For instance, EventNet\cite{sekikawa2019eventnet} adopts concepts from PointNet\cite{qi2017pointnet}, leveraging its architecture to effectively process the sparse and asynchronous nature of event data. Messikommer has borrowed the concept of SSC (Submanifold Sparse Convolutions)\cite{graham2017submanifold} from the field of 3D sparse point clouds and designed a sparse convolutional layer capable of handling asynchronous sparse event points\cite{Messikommer_2020_asynet}. This sparse convolutional layer uses a 'rulebook' to map input to output points. It engages only active points during sparse data processing, enhancing efficiency and memory usage. 

\begin{figure*}[!t]
	\centering
	\includegraphics[width=0.9\textwidth]{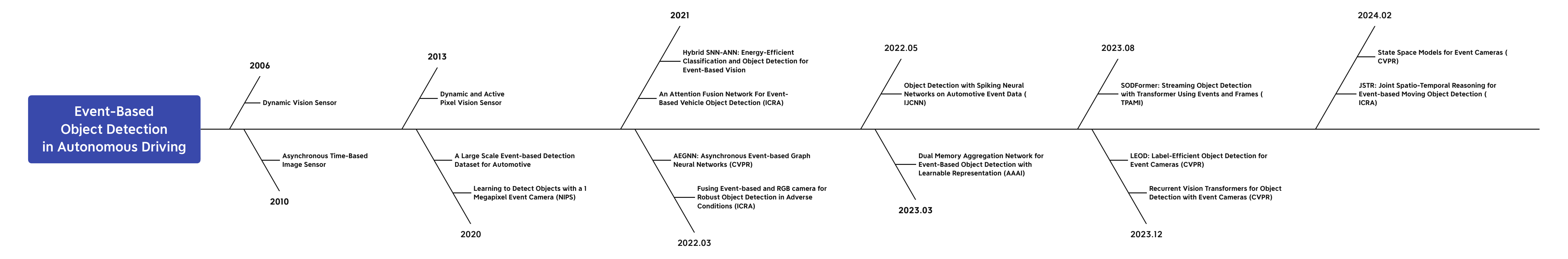}
	\caption{Publication dates of selected papers on event-based object detection in autonomous driving scenarios.}
	\label{fig:1}
\end{figure*}
\subsection{GNN for Event-Based Object Detection}
Treating event data from a fresh perspective involves viewing each event as a mutually connected node in the spatio-temporal domain. Graph neural networks (GNNs)\cite{GNN} utilize graph representations to model the characteristics of event streams. In this context, each event in the stream is considered a node in the graph, where the node's feature is the event's polarity. Considering a group of events, take event $e_i: (x_i,y_i,t_i,p_i)$ as a node $v_i$ in the graph. 

\begin{equation}
    d_{i,j}=\sqrt{\alpha(|x_i-x_j|^2+|y_i-y_j|^2)+\beta|t_i-t_j|^2}\leq R
\end{equation}

For every pair of nodes $v_i$ and $v_j$, if their spatio-temporal distance $d_{i,j}$ is less than a fixed radius distance $R$, an edge $e_{i,j}$ is connected between them. This process forms an undirected graph by connecting event nodes through edges based on their proximity in the spatio-temporal domain. Once the event stream is constructed as a graph, graph neural networks can be employed to process the event data, capturing the relationships among the event points. GNNs excel at processing graph-structured data, making them suitable for modeling the interconnected nature of events in event streams. By leveraging the power of GNNs, it becomes possible to extract meaningful representations that encode both the individual event properties and their relationships with other events.

Indeed, the dynamic nature of event data, where the graph needs to be continuously updated as events evolve in the temporal domain, can introduce significant computational overhead. To address this challenge, Schaefer proposed the Asynchronous Event-based Graph Neural Network (AEGNN)\cite{Schaefer_2022_GNN}. AEGNN treats event data as a temporally evolving graph, where each new event induces local changes to the GNN's activations, which are asynchronously propagated to lower layers.Unlike traditional graph neural networks, AEGNNs retain temporal information in the event data, rather than discarding it. This approach significantly reduces computation and latency because only the activations of nodes affected by each new event need to be recomputed, rather than recomputing all network activations. 

Building upon AEGNN, EAGR\cite{Gehrig_2022_EAGR} aimed to push the boundaries of graph neural network performance on event stream data. It achieved this by scaling up the model size by increasing the number of feature channels in the layer blocks and detection heads. Techniques like pruning node updates and early-time node aggregation are employed to reduce computation while maintaining performance.

\subsection{SNN for Event-Based Object Detection}
In embedded applications, algorithms have stringent requirements in terms of latency, accuracy, and power consumption. Traditional Deep Neural Networks (DNNs) face challenges in addressing these issues due to their typically continuous and synchronous computation modes, which consume significant energy and computational resources. By contrast, spiking neural networks (SNNs)\cite{SNN} are more biologically plausible, as neurons communicate through discrete and asynchronous spikes. This natural low-power and hardware-friendly mode of operation makes SNNs particularly suitable for sparse and binary event data, which encompasses spatial and temporal variations and can be accurately represented by spikes. Furthermore, SNNs enable the natural processing of temporal data without the need for additional complexity, as in recurrent neural networks (RNNs). This is because spike-based training enhances the ability to process spatio-temporal data, where the spatial aspect refers to neurons connecting only with nearby neurons to process input chunks separately, and the temporal aspect involves spike training occurring over time, allowing for the recovery of information lost in binary encoding through spike timing. Event cameras capture precisely this type of binary and sparse event data. Therefore, training SNNs directly on event data can lead to the design of fast and efficient embedded applications.

\subsubsection{Novel SNN Architectures}
Earlier studies have demonstrated the effectiveness of SNNs in event-based object recognition and image classification tasks\cite{gesture}\cite{Cordone_2022_IJCNN}, but their application in event-based object detection has been limited. However, in recent years, with the availability of large-scale event-based datasets, there has been a surge in research utilizing SNNs in the domain of event cameras.Building upon this foundation, Cordone\cite{Cordone_2022_spike_ssd} rebuilds effective object detection networks such as SqueezeNet\cite{iandola2016squeezenet}, VGG\cite{simonyan2015deep}, MobileNet\cite{howard2017mobilenets}, and DenseNet\cite{huang2018densely}, leveraging recent advancements in spike-based backpropagation techniques like surrogate gradient learning, parameterized Leaky Integrate-and-Fire (LIF) neurons, and the SpikingJelly framework\cite{doi:10.1126/sciadv.adi1480}. These advancements have enabled SNNs to achieve better performance when processing real-world event data. EMS-YOLO, on the other hand, aims to train deep SNNs for object detection tasks without converting them from ANNs\cite{su_2023_EMS_YOLO}. It introduces a full-spike residual block, EMS-ResNet, to extend the network depth efficiently with low power consumption. DOTIE\cite{Nagaraj_2023_DOTIE} introduces a more generalizable architecture for event-based object detection. It groups events using two dimensions, spatial and temporal, to separate events based on their originating objects and determine object boundaries. This asynchronous approach is robust to camera noise and performs well in scenarios where events are generated by stationary objects in the background. 

\subsubsection{SNN-ANN Hybrid}
Despite the advantages of SNNs in terms of low power consumption and their ability to directly process spatio-temporal information, their relatively short development history has resulted in limited performance and training challenges. To address these issues, many research efforts have focused on combining SNNs with traditional Artificial Neural Networks (ANNs), aiming to enhance SNN performance while maintaining low power consumption. Kugele\cite{Kugele_2021_hybrid} introduces a hybrid approach that combines SNNS and ANNs. This method utilizes event data as input, which is then processed and classified by the SNN. The hybrid approach aims to achieve higher processing capabilities while maintaining low energy consumption. 

\subsubsection{Spiking Neuron Enhancing}
Enhancing the intrinsic properties of SNNs, such as spike firing frequencies, and refining their training techniques are crucial for improving their performance in event-based object detection. Wang's approach\cite{Wang_2023_snn}, which differs from previous methods that relied on approximate gradient descent, utilizes Spike-Timing-Dependent Plasticity (STDP) for unsupervised training. On the other hand, FPDAG focuses on directly modulating the membrane potential dynamics of SNNs\cite{Zhang_2023_FPDAG}. By employing adaptive thresholds triggered by spikes, FPDAG stabilizes the training process and further boosts network performance. 

\subsection{Multi-Modal Fusion for Event-Based Object Detection}
Event cameras exhibit several advantages, including low power consumption, low latency, and high dynamic range. However, they encounter limitations when dealing with static scenes, as they are unable to perceive them effectively. Furthermore, event cameras lack the capability to capture texture information of objects, which can be crucial for accurate object recognition and detection. To overcome these limitations, the fusion of RGB image data and event data has been proposed as a solution. RGB images provide rich color and texture information, which is essential for recognizing and identifying objects. On the other hand, event data offers high sensitivity to motion, capturing fine details of moving objects. By combining these two modalities, one can leverage the complementary strengths of both data types, enabling effective perception and object detection in a wide range of scenarios, including low-light, high-light, and high-speed motion environments. The fusion of RGB and event data allows for the creation of a more comprehensive representation of the scene, taking into account both static and dynamic elements. By combining the rich texture information from RGB images with the motion sensitivity of event data, one can achieve more accurate and robust object detection in diverse conditions. This modality fusion approach has the potential to enhance the performance of perception systems in real-world applications, leveraging the unique advantages of both event cameras and traditional image sensors.

Tomy\cite{Tomy_2022_fusionFPN} utilized the FPN (Feature Pyramid Network)\cite{lin2017feature} to fuse event and image features extracted using ResNet\cite{resnet}. By integrating features from different levels, FPN enhances the detection of objects at various scales. JDF\cite{Li_2019_JDF} presents a joint detection framework that integrates event-based and frame-based perception for vehicle detection in autonomous driving. By fusing the two streams into a convolutional neural network (CNN) and utilizing a spiking neural network to generate visual attention maps, the framework achieves synchronization between the event and frame streams. Additionally, the paper introduces the Dempster-Shafer theory\cite{Liu2001_dempster} of evidence to fuse the outputs of the CNN, thereby enhancing the performance of vehicle detection. Liu introduced an attention mechanism\cite{Liu_2021_ADF}, designing an Attention-based fusion module. This module uses squeeze-and-excitation blocks and channel attention mechanisms to deeply fuse the feature maps of Gaussian parameters and confidence maps. This approach allows for capturing richer feature information and improving the accuracy of vehicle object detection. 

However, the aforementioned studies are based on the fusion of RGB images and event data within a single timeframe and do not consider the rich semantic information of event data along the temporal dimension. Therefore, RENet has designed a Temporal Multi-scale Aggregation Module to fully exploit information from both RGB exposure times and widely spaced event frames. Additionally, a Bidirectional Fusion Module is introduced to calibrate and integrate multimodal features. SODFormer\cite{Li_2023_sod} takes a further step by improving upon Deformable DETR\cite{zhu2021deformable}, completely abandoning the CNN structure. Instead, it relies entirely on attention mechanisms for frame data processing, event stream processing, and data fusion. By aggregating RGB and frame images over a period of time on the temporal scale using Temporal Deformable DETR\cite{zhou2022transvod}, SODFormer is capable of fully leveraging the information along the temporal dimension. This approach allows for more efficient and accurate feature extraction and fusion, leading to improved object detection performance. 
\begin{table*}[!t]
        \caption{A comparative analysis of open-source datasets for event-based object detection. The DSEC-Det dataset is derived from the DSEC dataset\cite{DSEC}. It utilizes qdtrack\cite{qdtrack_conf} for detection on RGB images, followed by manual rectification to generate the final dataset. The event camera resolution for DSEC-Det is 640$\times$480, while the RGB camera boasts a resolution of 1280$\times$720.}
	\label{tab:table2}
	\vspace{-0.35cm}
	\begin{center}
		\setlength{\tabcolsep}{1.60mm}{
			\begin{tabular}{lcccccccc}
				\toprule
				Dataset & Year  & Venue    & Resolution & Modality  & Classes  & Boxes & Label & Frequency   \\
				\hline
                    Gen1 Detection~\cite{de_Tournemire_2020_red} & 2020  & arXiv   & 304$\times$240 & Events  & 2  & 255k & Pseudo & 1, 4 Hz  \\
				1 Mpx Detection~\cite{Perot_2020_dataset} &  2020 & NIPS  & 1280$\times$720 & Events   & 3  & 25M & Pseudo & 60 Hz  \\
                    \hline
				PKU-DDD17-CAR~\cite{Li_2019_JDF} & 2019 & ICME   & 346$\times$260 & Events, Frames   & 1  & 3155 & Manual  & 1 Hz  \\
                    DSEC-Det~\cite{DSEC} & 2022 & RA-L & 640$\times$480, 1280$\times$720 & Events, Frames & 8 & 390.1k & Pseudo & 20 Hz \\
				PKU-DAVIS-SOD~\cite{Li_2023_sod} & 2023  & TPAMI  & 346$\times$260 & Events, Frames  & 3  & 1080.1k & Manual & 25 Hz  \\
				\bottomrule
		\end{tabular}}
	\end{center}
	\vspace{-0.30cm}
\end{table*}

\section{Event Object Detection Dataset}
To support the rapid development of event-based object detection in recent years, several datasets have been proposed. Unlike traditional image-based object detection that identifies target objects on individual images, event-based object detection more closely resembles the detection of objects in videos with continuous semantic information along the temporal dimension. To fully leverage the rich spatio-temporal semantic information provided by the mechanism of event cameras, datasets proposed in recent years often involve collecting continuous scene data using event cameras over a period of time. While there are examples of transforming classic image object detection datasets into event data for object detection, this approach loses the characteristic of spatio-temporal continuity of event data, and thus is not within the scope of this paper. 

Most publicly available datasets focus on the field of object detection in autonomous driving scenarios. The requirements for low latency, high dynamic range, and low power consumption in vehicle-mounted sensors for autonomous driving make event cameras a strong contender to replace traditional vehicular visual sensors. This has also led to a preference among researchers for object detection in autonomous driving scenarios. These datasets can generally be divided into two categories: one consisting of pure event data, and the other containing both traditional camera and event camera data. The data is usually vast in volume, and a detailed comparison table can be found in Table 1.

\subsection{Event Only Datasets} 
The Gen1 dataset\cite{de_Tournemire_2020_red} is the first large-scale object detection dataset that does not rely on video conversion but instead directly utilizes event cameras for real-world data acquisition. By aggregating and transforming the asynchronous intensity measurements from the ATIS sensor into grayscale images, it enables humans to annotate the event stream. However, the Gen1 dataset employs a GEN1 event sensor with a relatively low resolution. Consequently, the 1 Mpx dataset\cite{Perot_2020_dataset} utilizes a higher-resolution event camera. In order to reduce the annotation costs of the dataset, researchers placed a traditional camera alongside the event camera. They annotated the RGB video using commercial automatic annotation software and then mapped the automatically annotated calibration boxes from the frame images to the event stream.

\subsection{Multi-modal Datasets} 
Although event cameras can output rich semantic information, they may malfunction under certain conditions, such as when there is no movement or during low-speed scenarios. To address this issue, the introduction of semantic information from other modalities can effectively compensate for this deficiency of event cameras. With the development of event camera sensors, the advent of DAVIS (Dynamic and Active-Pixel Image Sensors) enables the simultaneous recording of event signals and light signals at the same pixel point, thereby outputting both RGB data and event data concurrently. 

The DDD17 dataset\cite{binas2017ddd17} is the first to use DAVIS to record image data and event data simultaneously. Building upon this, the PKU-DDD17 dataset\cite{Li_2019_JDF} introduces a multimodal dataset that includes both image and event data, manually annotated on RGB images. Since DAVIS captures both modalities from the same pixel point, this dataset allows users to directly integrate the two modalities of data. Similarly, the PKU-SOD dataset\cite{Li_2023_sod}, through the collection of data under scenarios where traditional cameras fail, compares the two modalities of data manually and extensively examines conditions under low light and high speed. 

Although DAVIS can output data from both modalities without any adjustments, the resolution of its event and image data is low, and the image quality is relatively poor. One solution is to use a higher-resolution event camera in conjunction with a traditional RGB camera to record frame data. Inspired by KITTI\cite{KITTI} and MVSEC\cite{MVSEC}, the DSEC dataset\cite{DSEC} employs a stereoscopic multimodal camera to record data from autonomous driving scenarios, along with the vehicle's GPS data and depth map data. Subsequently, the DSEC-Det dataset utilizes existing traditional RGB frame object detection algorithms to obtain bounding boxes, supplemented by manual corrections. As DSEC records data from traditional cameras and event cameras of the surrounding scene at the same moment, it is necessary to rectify the image data when using it. Although it cannot output two modalities of data with the same pixel coordinates directly like DAVIS, the independent use of traditional cameras and event cameras brings higher resolution to both image and event data.

\section{Challenges}
The field of object detection based on event cameras is still in its infancy, with a significant gap compared to traditional frame-based object detection. Although event data exhibits stronger semantic continuity in the spatio-temporal dimensions, the asynchronous nature of such data remains challenging to process. Methods that compress events within a fixed time frame simplify the asynchronous aspect of event data but at the cost of losing a wealth of temporal information. Employing Spiking Neural Networks (SNNs), which share the same asynchronous nature as event data, seems to be an ideal approach. This is because the spike information in SNNs is fundamentally similar to event data, both being simulations of biological signals. However, SNNs are still in the early stages of development. While they have shown promising performance in classification tasks, their performance in regression problems is still not on par with traditional deep neural networks. Additionally, the use of SNNs often involves specialized hardware architectures, and the software ecosystem for SNNs is yet to mature.

Moreover, as mentioned in the previous context regarding the failure of event cameras in static conditions, the integration of modalities using conventional cameras is a potential solution. However, there is still room for improvement in terms of real-time performance. Overall, the enhancement of object detection capabilities based on event cameras is not solely dependent on algorithmic advancements. It also necessitates improvements in hardware performance and may even be related to the development of brain-inspired intelligence. The advancement of this technology could potentially benefit from interdisciplinary research that combines insights from computer vision, neuroscience, and hardware engineering to overcome current limitations and fully leverage the unique advantages of event-based data.

\section{Conclusion}
The utilization of event cameras for object detection, despite being in its nascent stage, has garnered significant interest in recent years. This is largely attributable to the inherent characteristics of event cameras, which include low power consumption, high dynamic range, and low latency. Additionally, the biological significance of these cameras—mimicking the asynchronous, event-driven nature of human vision—makes them a compelling choice for various applications. In this article, we have summarized several types of deep learning methodologies currently employed for object detection using event cameras and introduced the corresponding datasets. Our aim is to assist researchers and engineers in taking the first step towards further exploration and development of event-based object detection tasks.

\end{document}